\documentclass[10pt,twocolumn,letterpaper]{article}

\usepackage{wacv}      

\usepackage{graphicx}
\usepackage{amsmath}
\usepackage{amssymb}
\usepackage{booktabs}
\usepackage{enumitem}
\usepackage{booktabs}
\usepackage{siunitx}
\usepackage[table]{xcolor}
\usepackage{tabularx}
\usepackage{caption}
\usepackage{xcolor}
\usepackage{adjustbox}
\usepackage{parskip}
\newif\iffigures

\usepackage[pagebackref,breaklinks,colorlinks]{hyperref}
\newcommand{\textbfgreen}[1]{\textbf{\textcolor{green!80!black}{#1}}}
\newcommand{\textbfred}[1]{\textbf{\textcolor{red!70!black}{#1}}}
\newcommand{\textbfblue}[1]{\textbf{\textcolor{blue!70!black}{#1}}}
\usepackage{multirow}
\usepackage[capitalize]{cleveref}
\crefname{section}{Sec.}{Secs.}
\Crefname{section}{Section}{Sections}
\Crefname{table}{Table}{Tables}
\crefname{table}{Tab.}{Tabs.}

\setlist[itemize]{itemsep=0.04cm}
\def\wacvPaperID{*****} 
\def\confName{WACV}
\def\confYear{2024}

\begin{document}

\title{SynergyNet: Bridging the Gap between Discrete and Continuous Representations for Precise Medical Image Segmentation}

\author{Vandan Gorade$^1$, Sparsh Mittal$^1$, Debesh Jha$^2$, Ulas Bagci$^2$\\
$^1$ Indian Institute of Technology Roorkee, India\\
$^2$ Machine \& Hybrid Intelligence Lab, Department of Radiology, Northwestern University, USA}

\maketitle

\begin{abstract}
In recent years, continuous latent space (CLS) and discrete latent space (DLS) deep learning models have been proposed for medical image analysis for improved performance. However, these models encounter distinct challenges. CLS models capture intricate details but often lack interpretability in terms of structural representation and robustness due to their emphasis on low-level features. Conversely, DLS models offer interpretability, robustness, and the ability to capture coarse-grained information thanks to their structured latent space. However, DLS models have limited efficacy in capturing fine-grained details. To address the limitations of both DLS and CLS models, we propose \textit{SynergyNet}, a novel bottleneck architecture designed to enhance existing encoder-decoder segmentation frameworks. SynergyNet seamlessly integrates discrete and continuous representations to harness complementary information and successfully preserves both fine and coarse-grained details in the learned representations. Our extensive experiment on multi-organ segmentation and cardiac datasets demonstrates that SynergyNet outperforms other state of the art methods including TransUNet: dice scores improving by 2.16\%, and Hausdorff scores improving by 11.13\%, respectively. When evaluating skin lesion and brain tumor segmentation datasets, we observe a remarkable improvements of 1.71\% in Intersection-over-Union scores for skin lesion segmentation and of 8.58\% for brain tumor segmentation. Our innovative approach paves the way for enhancing the overall performance and capabilities of deep learning models in the critical domain of medical image analysis.

\end{abstract}


\vspace{-3mm}
\section{Introduction}
Medical image segmentation, a key step in gaining vital anatomical insights, assists clinicians in injury identification, disease monitoring, and treatment planning. As reliance on medical image analysis grows, the demand for precise, robust segmentation techniques rises. In this regard, deep learning has greatly improved our ability to do this. Existing deep learning models can be divided into continuous latent space (CLS) and discrete latent space (DLS) models. The CLS models represent latent variables as continuous values, enabling fine-grained representation.  

CLS Models such as FCNs~\cite{long2015fully}, UNet~\cite{ronneberger2015u}, and TransUNet~\cite{chen2021TransUNet}  and others~\cite{oktay2018attention, cao2022swin,heidari2023hiformer} have shown an ability to capture spatial relationships and fine-grained details for medical image segmentation. However, these models offer limited latent interpretable representations of structural information and robustness~\cite{santhirasekaram2022vector} in terms of generalization. DLS methods employ discrete codes instead of continuous values for latent variables. They use techniques such as vector quantization to discretize the latent space into a finite set of elements representing anatomical structures. This enables efficient and generalized data representation. Approaches such as VQVAE~\cite{van2017neural} and VQGAN~\cite{esser2021taming} have shown promise in image generation, representation learning, and data compression.

Recent studies~\cite{pinaya2022unsupervised,gangloff2022leveraging,santhirasekaram2022vector,graham2022transformer,jin2022deep} highlight the effectiveness of DLS models in achieving interpretable and robust medical segmentation, particularly for organs like lungs, retinas, optic discs, and prostates. However, DLS models struggle to capture fine-grained details and complex spatial relationships, especially in multi-organ and cardiac segmentation tasks. Accurate modeling of spatial interdependencies between organs is crucial for precisely segmenting intricate boundaries and overlapping structures. Recent studies ~\cite{radford2021learning,li2021align,nagrani2021attention,farshad2022net,chen2021TransUNet,yuan2023effective,raghu2021vision,wang2022rethinking} have highlighted the advantages of learning complementary information across various domains, including medical imaging. Motivated by this trend, our study aims to address the pivotal question: \textit{``\textbf{How can we effectively integrate complementary information from discrete and continuous latent space models for improved medical image segmentation?}''}.

We present \textit{SynergyNet}, a novel bottleneck architecture designed specifically for encoder-decoder segmentation models, aiming to enhance medical image segmentation results by integrating continuous and discrete latent spaces. SynergyNet includes the Quantizer, DisConX, and Refinement modules. The encoder extracts a detailed continuous representation, while the quantizer module maps it to a compact discrete representation using vector quantization. By reducing dimensionality, the quantizer module enables efficient, structured representation while preserving essential information. The DisConX module serves as a bridge, employing cross-attention to effectively combine the discrete and continuous representations. Leveraging their complementary information, the DisConX module enhances pattern capture and interpretation. The refinement module further enhances the fused features, using hard attention to emphasize essential elements and filter out noise. The refinement module improves discriminative power and segmentation quality by focusing on relevant features. Our contributions are as follows:

\begin{itemize}
    \vspace{-1.5mm}
    \item We propose SynergyNet, a novel method that integrates discrete and continuous representations to enhance medical image segmentation performance. This integration has not been explored in prior studies for medical image segmentation tasks. 
    \vspace{-1.5mm}
    \item Our study demonstrates the effectiveness of combining CLS and DLS models in improving model generalization across diverse datasets. By leveraging CLS models for fine-grained detail capture and DLS models' structured latent space for encoding coarse-grained details, we observe notable enhancements in learning and generalization. This integration effectively utilizes the strengths of each approach, resulting in improved performance across various datasets. 
    \vspace{-1.5mm}
    

    \item SynergyNet is extensively evaluated on four diverse datasets, including Synapse multi-organ segmentation, ACDC dataset for cardiac segmentation, ISIC 2018 dataset for skin lesion segmentation, and brain tumor segmentation dataset. Results show that SynergyNet outperforms both CLS~\cite{ronneberger2015u,oktay2018attention,dosovitskiy2020image,chen2021TransUNet} and DLS-based methods~\cite{van2017neural} across all evaluated datasets. Qualitative analysis confirms the efficacy of SynergyNet in capturing intricate anatomical structures and achieving more precise segmentation compared to existing methods
    
\end{itemize}



\begin{figure*}[!t]
\centering
\def\svgwidth{\columnwidth}
\includegraphics[width=2.2\columnwidth,scale=1.9]{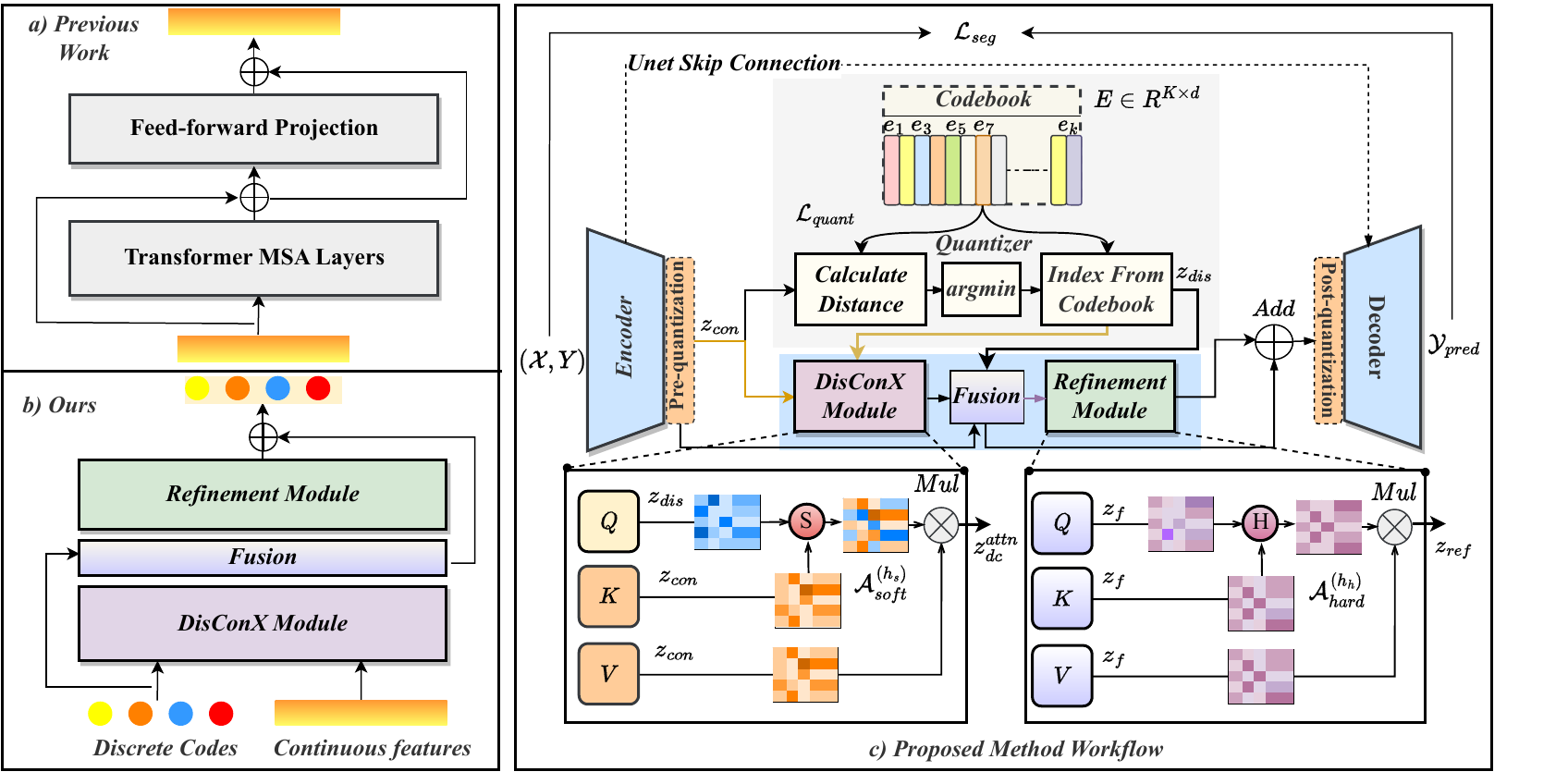}
\caption{
Figures (a) and (b) compare the bottleneck architecture of our proposed SynergyNet with the existing work ~\cite{chen2021TransUNet}. Figure (c) illustrates the workflow of our architecture, where the input image is encoded to generate a continuous representation $z_{con}$. This continuous representation is quantized to obtain discrete codes $e_k$ and forms a discrete representation $z_q$. The DisConX module (Section~\ref{subsection:disconx}) is applied to both the continuous and discrete representations, resulting in a synergized representation $z^{attn}_{dc}$. The continuous representation $z_{con}$, discrete representation $z_q$, and attended representation $z^{attn}_{dc}$ are then fused and processed through the refinement module (Section~\ref{subsection:ref}). The output of the refinement module is passed to the decoder to obtain the final result.}
\label{fig: workflow_1}
\end{figure*}

\section{Proposed Method}
\label{sec:method}
We first discuss the preliminaries (Section~\ref{section:pre}) and then present our newly proposed algorithm (SynergyNet) and its architecture (Section~\ref{section:arch}). 

\subsection{Preliminaries}\label{section:pre}
\subsubsection{Problem Statement} \label{subsection:ps}

Medical image segmentation aims to automatically label anatomical structures or pathological regions within medical images. Mathematically, this involves finding a mapping function $f$ that assigns labels $y$ to pixels $x$ in the input image domain $\mathcal{X}$. The goal is to maximize the conditional probability of the ground truth segmentation labels $\hat{y}$ given the input image $x$, i.e., $\hat{y} = \arg\max_{y} P(y|x)$. Learning the parameters of the mapping function $f$ involves assigning the correct labels to each pixel using training data. The learning process employs a loss function that usually consists of Binary Cross Entropy (BCE) and Dice similarity coefficient. This loss function can be defined as follows:
\vspace{-1.5mm}
\begin{equation}
\label{eq:1}
L_{seg} = {BCE}(y, \hat{y}) + (1 - {Dice}(y, \hat{y})),
\end{equation}
where ${BCE}(y, \hat{y})$ calculates the binary cross entropy loss between the predicted labels $y$ and the ground truth segmentation $\hat{y}$, and ${Dice}(y, \hat{y})$ computes the dice similarity coefficient between $y$ and $\hat{y}$. 
\vspace{-2.6mm}
\subsubsection{Vector Quantization} 
\label{subsection:vq}

Following VQVAE\cite{van2017neural}, Vector quantization (VQ) transforms continuous latent space vectors $z_{con} \in \mathbb{R}^{dim}$ into discrete codes $e_k$ from a predefined codebook $E \in \mathbb{R}^{K \times dim}$, where $K$ is the codebook size. The objective of VQ is to find the code $e_k$ from the codebook that minimizes the euclidean distance to the input vector $z_{con}$. This code $e_k$ serves as the discrete representation $z_{dis}$ of $z_{con}$. During training, the codebook $E$ and the mapping functions between the continuous and discrete representations are learned by minimizing the quantization loss $\mathcal{L}_{quant} = \lVert z_{con} - e_k \rVert_2^2$. The quantization process efficiently encodes and decodes data while preserving important information in discrete representations. We use the total loss function $   \mathcal{L}_{total} = \mathcal{L}_{seg} + \mathcal{L}_{quant}$ for end-to-end model training. 
\vspace{-2.6mm}
\subsubsection{Multi-head Cross-attention Mechanism} 
\label{subsection:multihead_ca}
The multi-head cross-attention mechanism extends the cross-attention by incorporating multiple attention heads. Each attention head attends to different subspaces of queries and keys, capturing diverse relationships and dependencies. Given a set of queries $Q$ and keys $K$, multiple sets of attention weights are computed, one for each attention head. The relevance scores between a query $q_i$ and a key $k_j$ are obtained using a similarity function denoted as ${score}(q_i, k_j) = {sim}(q_i, k_j)$. The softmax function is applied to transform the relevance scores into attention weights for each attention head:
\begin{equation}
\mathcal{A}_{soft}^{(h)}(q_i, k_j) = \frac{\exp({score}^{(h)}(q_i, k_j))}{\sum_{j'} \exp({score}^{(h)}(q_i, k_{j'}))}
\end{equation}
The multi-head cross-attention mechanism then computes a weighted sum of the values associated with the keys using the attention weights of each attention head $h$:
\begin{equation}
z^{attn}_h = \sum_j \mathcal{A}_{soft}^{(h)}(q_i, k_j) \cdot v_j.
\end{equation}
Here, $z^{attn}_h$ represents the aggregated result for the given query $q_i$, considering the importance assigned by the attention weights of the $h$-th attention head. The outputs from all the attention heads are concatenated and linearly transformed to produce the final output:
\begin{equation}
\label{eq:attn_cat}
z^{attn} = Concat(z^{attn}_1, z^{attn}_2, \ldots, z^{attn}_h) \cdot W^O.
\end{equation}
The multi-head cross-attention mechanism enables the model to capture various interactions and dependencies between queries and keys, enhancing its representation and information retrieval capabilities.

\subsection{Proposed Architecture} 
\label{section:arch} 
Our proposed architecture, illustrated in Fig.~\ref{fig: workflow_1}(c), consists of three key components: the encoder, bottleneck, and decoder. The bottleneck incorporates the \textit{Quantizer, DisConX,} and \textit{Refinement }modules. Starting with an input image $X$, the encoder function $f$ generates the continuous representation $z_{con}$. The Quantizer module (Section~\ref{subsection:vq}) maps $z_{con}$ to a more compact discrete representation $z_{dis}$, capturing essential information efficiently. The DisConX module (Section~\ref{subsection:disconx}) combines the discrete and continuous representations through cross-attention, leveraging both benefits to enhance data interpretation. The Refinement module (Section~\ref{subsection:ref}) further improves the representation by emphasizing relevant features. This step enhances the model's discriminative power for the given task.
\vspace{-2.6mm}
\subsubsection{DisConX Module} \label{subsection:disconx}
The DisConX module integrates the discrete representation $z_{{dis}}$ and continuous representation $z_{{con}}$ using cross-attention, as discussed in Section~\ref{subsection:multihead_ca}. It calculates relevance scores between discrete queries $q_i$ and continuous keys $k_j$, and computes attention weights using a softmax function. The module then performs a weighted sum of the continuous values $v_j$ associated with the keys based on the attention weights. The computation happens as: 
\begin{equation}
z^{attn}_{dc}= \sum_{j} \mathcal{A}^{(h_{s})}_{soft}(z_{{dis}}, z_{{con}}) \cdot z_{{con}}.
\vspace{-2.5mm}
\end{equation}

Here, $h_{s}$ represents the index of the attention heads, $z_{{dis}}$ denotes the discrete query, and $z_{{con}}$ represents the continuous key/value. The resulting $z^{attn}_{dc}$ is the aggregated representation, considering the attention weights from all attention heads. This integration of discrete and continuous representations enables the exchange of complementary information, enhancing the model's ability to capture complex patterns and improving performance in tasks such as semantic segmentation. 
Next, the information of $z_{{dis}}$, $z_{{con}}$, and $z_{{attn}}^{{dc}}$ is fused  as $z_{f} = {Fusion}(z_{{dis}}, z_{{con}}, z_{{dc}}^{{attn}})$. 

The fusion operation integrates complementary information from discrete and continuous representations, enhancing the overall representation for subsequent refinement modules. We empirically choose addition for fusion.
\vspace{-2.5mm}
\subsubsection{Refinement Module}
\label{subsection:ref}
The proposed refinement module incorporates a hardness-aware self-attention mechanism, which captures the relevance and similarity between elements in the fused representation. This mechanism enhances the overall representation quality by emphasizing important elements and filtering out the noise. The element with the highest relevance score is identified as the most important. The attention weight for each element is determined by comparing its similarity to other elements. The equation below represents this calculation:
\begin{equation}
\mathcal{A}^{(h_{h})}_{hard}(z_{{f}_i}) = \mathbb{I}({sim}^{(h_h)}(z_{{f}_i}, z_{{f}j}) = \max_j {sim}^{(h_h)}(z_{{f}_i}, z_{{f}_j})).
\end{equation}
Here, the indicator function $\mathbb{I}$ checks if the similarity between element $z_{{f}_i}$ and any other element $z_{{f}_j}$ is the maximum among all similarities. $h_h$ is the index of attention head. 
Next, the self-attention mechanism calculates a weighted sum of the values associated with the selected elements using the attention weights for each attention head:
\begin{equation}
z_{{ref}} = \sum_j \mathcal{A}^{(h_{h})}_{hard}(z_{{f}_i}) \cdot (z_{{f}_j}).
\vspace{-2.5mm}
\end{equation}
The refined information $z_{{ref}}$ represents the output of the self-attention mechanism for the $h_h$-th attention head. This process is repeated for all attention heads. The resulting refined information from all the attention heads is then concatenated and linearly transformed to produce the final refined representation. It highlights the most important elements within the fused representation, considering multiple attention heads. This refined representation enhances the discriminative power and overall quality of the fused features. Finally, the fused representation $z_{\text{f}}$ is added to $z_{ref}$ and then passed through the decoder.



\begin{table*}[!ht]
\centering
\caption{Quantitative results for multi-organ segmentation: \textbfgreen{Green} - best, \textbfblue{Blue} - second-best (R50=ResNet50). }
\label{tab:synapse}
\adjustbox{width=1.0\textwidth}{
\begin{tabular}{lS[table-format=2.2]S[table-format=2.2]|S[table-format=2.2]S[table-format=2.2]S[table-format=2.2]S[table-format=2.2]S[table-format=2.2]S[table-format=2.2]S[table-format=2.2]S[table-format=2.2]}
\toprule
\textbf{Method} & \multicolumn{2}{c}{\textbf{Mean scores}} & \multicolumn{8}{c}{\textbf{Organ-wise dice similarity coefficient (DSC)}} \\
\cmidrule(lr){2-3} \cmidrule(lr){4-11}
 & {\textbf{DSC}} & {\textbf{HD}} & \multicolumn{1}{c}{\textbf{Aorta}} & \multicolumn{1}{c}{\textbf{Gallbladder}} & \multicolumn{1}{c}{\textbf{KidneyL}} & \multicolumn{1}{c}{\textbf{KidneyR}} & \multicolumn{1}{c}{\textbf{Liver}} & \multicolumn{1}{c}{\textbf{Pancreas}} & \multicolumn{1}{c}{\textbf{Spleen}} & \multicolumn{1}{c}{\textbf{Stomach}} \\
\midrule


R50 UNet        & 74.68  & 36.87  & 87.74  & 63.66  & 80.60  & 78.19  & 93.74  & 56.90  & 85.87  & 74.16 \\

DeepLabV3+ & 77.63 & 39.95 & \color{green}{\textbf{88.04}} & 66.51 & \color{blue}{\textbf{82.76}} & 74.21 & 91.23 & 58.27 & 87.43 & 73.53 \\

R50 Att-UNet     & 75.57  & 36.97  & {55.92}  & 63.91  & 79.20  & 72.71  & 93.56  & 49.37  & 87.19  & 74.95 \\
R50 ViT          & 71.29  & 32.87  & 73.73  & 55.13  & 75.80  & 72.20  & 91.51  & 45.99  & 81.99  & 73.95 \\
TransUNet        & {77.48}  & \textbfblue{31.69}  & \textbfblue{87.23}  & \textbfblue{63.13}  & 81.87  & 77.02  & 94.08  & 55.86  & 85.08  & 75.62 \\
\midrule
VQUNet            & {63.44}  & {68.79}  & 78.99  & 50.74  & {67.32}  & 61.91  & {89.94}  & {33.96}  & {73.83}  & {50.87}  \\
TransVQUNet-2s2h    & 65.44  & 40.79  & 80.95  & {48.29}  & 71.42  & {61.89}  & 90.90  & 36.88  & 77.59  & 55.61  \\
TransVQUNet-8s2h    & 68.41  & 35.05  & 83.57  & 54.53  & 73.73  & 66.21  & 92.39  & 39.45  & 80.26  & 57.10  \\
\midrule
SynergyNet-2s2h      & \textbfblue{78.81}  & 26.19  & 85.31  & 61.14  & {81.89}  & \textbfgreen{79.75}  & \textbfblue{94.42}  & {56.66}  & {89.81}  & \textbfgreen{81.51}  \\
SynergyNet-8s2h     & \textbfgreen{79.65}  & {23.29}  & {86.10}  & \textbfgreen{65.49}  & \textbfgreen{82.78}  & \textbfblue{79.23}  & \textbfgreen{95.06}  & \textbfblue{58.28}  & \textbfblue{88.95}  & \textbfblue{81.30}  \\
SynergyNet-8s8h     & 
{77.33}  & \textbfgreen{20.56}  & 85.79 & 
61.11 & 
81.69 & 
77.07 & 
94.44 & 
\textbfgreen{64.80} & 
86.40 & 
75.28 \\
\bottomrule
\end{tabular}}
\end{table*}

\section{Experimental Platform}\label{sec:platform}
\textbf{Datasets:} We utilized four open-source medical segmentation datasets for our experiments. The Synapse Multi-Organ Segmentation dataset~\cite{synapse} consists of 30 clinical CT cases with annotated segmentation masks for eight abdominal organs. We followed the configuration described in~\cite{chen2021TransUNet}, using 18 cases for training and 12 cases for testing. The ACDC dataset~\cite{acdc} is a cardiac MRI dataset with 100 exams, including labels for the left ventricle (LV), right ventricle (RV), and myocardium (MYO). We divided the dataset into 70 training samples, 10 validation samples, and 20 testing samples as per~\cite{chen2021TransUNet}. For skin lesion segmentation, we adopted the ISIC 2018 dataset~\cite{isic18} and followed the division into train, validation, and test sets as per previous work \cite{azad2019bi,ruan2022malunet}. The Brain Tumour Segmentation (BTS) dataset~\cite{bts} comprises 233 volumetric T1-weighted contrast-enhanced images from 233 patients (with a total of 3064 2D slices), including three types of brain tumors (meningioma, glioma, and pituitary tumor) with corresponding binary masks. We maintained an approximate 80:20 ratio for the training and test sets.

\textbf{Metrics:}  We utilize the Dice Similarity Coefficient (DSC) and the 95\% Hausdorff Distance (HD) metrics for the synapse and ACDC datasets to follow the segmentation challange standards and benchmarking. For the ISIC-18 and BTS datasets, we employ a more comprehensive range of metrics per segmentation challenge benchmarking, including the Intersection over Union (IOU), DSC, Specificity (SP), Sensitivity (SE), and Accuracy (ACC). For HD, lower is better. For other metrics, higher is better. 

\textbf{Implementation Details:}
We use PyTorch framework and train the models on three RTX 2080 GPUs, each with 11GB of memory. The input image size was set to 224 × 224. During training, we used a batch size of 8 and a learning rate of 0.01. We utilized the SGD optimizer with a momentum of 0.9 and weight decay of 0.0001. We employed data augmentations, such as flipping and rotating.

\textbf{Architecture Configuration:}
SynergyNet employs a ResNet50 encoder pre-trained on the ImageNet dataset, although we have no restriction on the choice of architecture for encoder. The quantizer module utilizes a codebook size of $K=1024$. The quantizer, DisConX and Refinement module maintain a hidden dimension of $dim=512$. We evaluate multiple SynergyNet variants, for example, SynergyNet-8s2h implies that $h_s=8$ and $h_h=2$, i.e., it has 8 DisConX heads and 2 refinement heads. The pre-and post-quantization blocks consist of two convolution block. The decoder has the same depth as the encoder.

\textbf{Techniques for Comparison:}  We compare SynergyNet against four CLS methods, i.e., UNet~\cite{ronneberger2015u}, Att-UNet~\cite{oktay2018attention}, DeeplabV3+~\cite{chen2018encoder} R50ViT~\cite{dosovitskiy2020image,chen2021TransUNet}, TransUNet~\cite{chen2021TransUNet} and two DLS methods, i.e.,VQUNet~\cite{van2017neural,Szymanowicz_2022_WACV,santhirasekaram2022vector} and TransVQUNet. TransVQUNet architecture is a combination of VQUNet and TransUNet. It consists of an encoder followed by a quantizer module and a transformer bottleneck, similar to the bottleneck of TransUNet. We kept the hyperparameters and architectural design consistent across all the methods for consistency. 
 


\section{Experimental Results}\label{sec:results}
\subsection{Synapse multi-Organ segmentation}



\begin{figure*}[!t]\centering
\def\svgwidth{\columnwidth}
\includegraphics[width=2.0\columnwidth,scale=1.8]{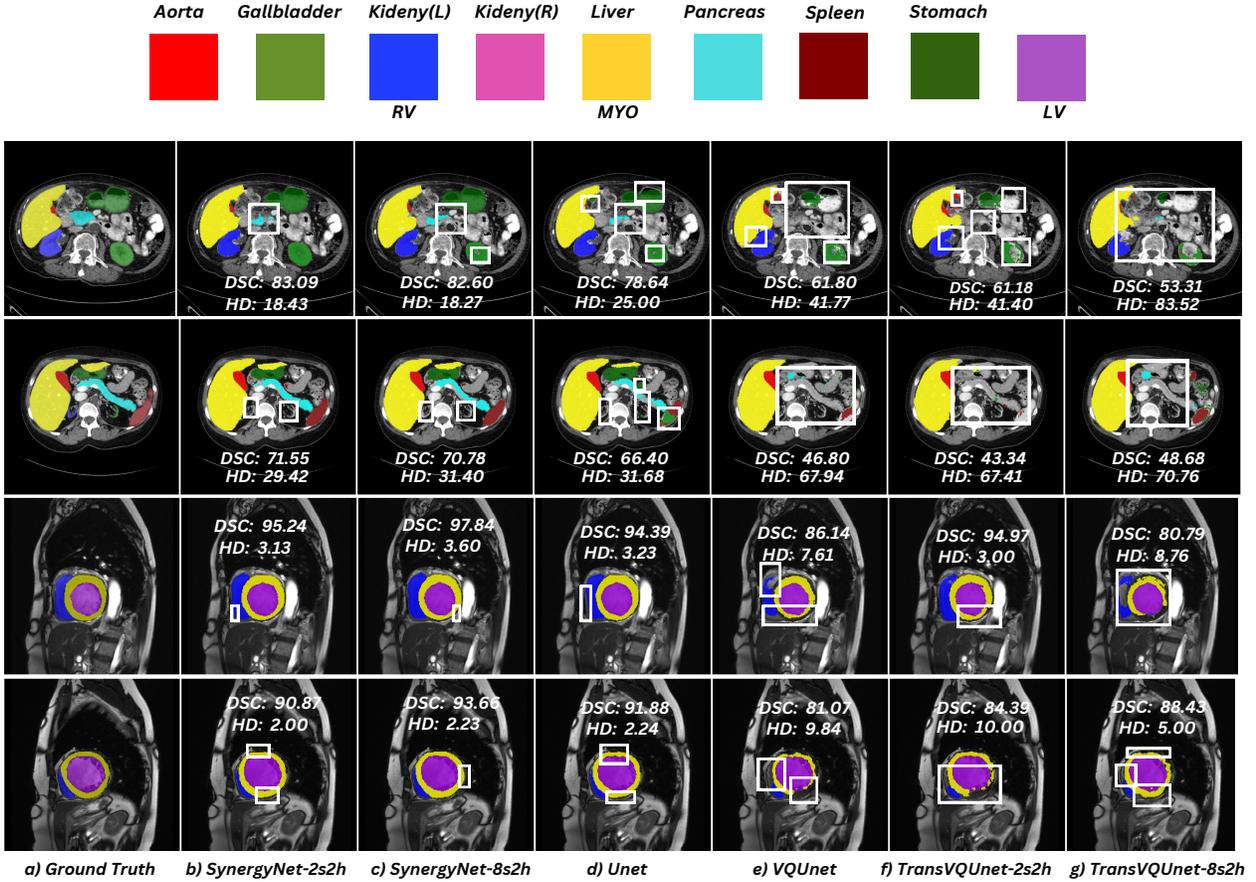}
\caption{Delineations on Synapse (first and second row) and ACDC (third and fourth row) datasets are shown with color-coded (First row, yellow: liver, blue: right kidney, green: left kidney, light blue: pancreas. Second row, blue, purple, and yellow represent the RV, LV, and MYO, respectively.). The overlapping white bounding box represents errors made by models.}
\label{pred:synaps_qualitative}
\vspace{-4mm}
\end{figure*}

Table~\ref{tab:synapse} compares SynergyNet with both CLS and DLS methods. SynergyNet outperforms both CLS and DLS methods by a significant margin. Quantitatively, SynergyNet-8s2h achieves a 2.17pp improvement in DSC and a 12.20pp deterioration in HD compared to TransUNet, while showing an 11.24pp improvement in DSC and an 11.16pp deterioration in HD compared to TransVQUNet-8h (pp= percentage point). SynergyNet-8s8h variant shows the best results in terms of  HD metric. SynergyNet demonstrates superior accuracy in delineating the organs and capturing the boundary between them.  
It outperforms other methods in learning both coarse-grained anatomical structures (e.g., stomach and liver) and fine-grained anatomical structures (e.g., gallbladder and spleen). TransUNet, a well-engineered CLS method, exhibits comparable performance in learning fine-grained structures. On the other hand, DLS methods can capture coarse anatomical structures but struggle to capture fine-grained boundaries. SynergyNet benefits from the complementary information extracted by continuous and discrete latent spaces. Fig.~\ref{pred:synaps_qualitative} further highlights the effectiveness of SynergyNet in accurately segmenting fine/coarse and complex structures. SynergyNet yields more robust and precise segmentation results even in the presence of intricate variations.
\vspace{-3mm}

\begin{figure}[htbp]\centering
\def\svgwidth{\columnwidth}
\includegraphics[width=0.9\columnwidth,scale=1.1]{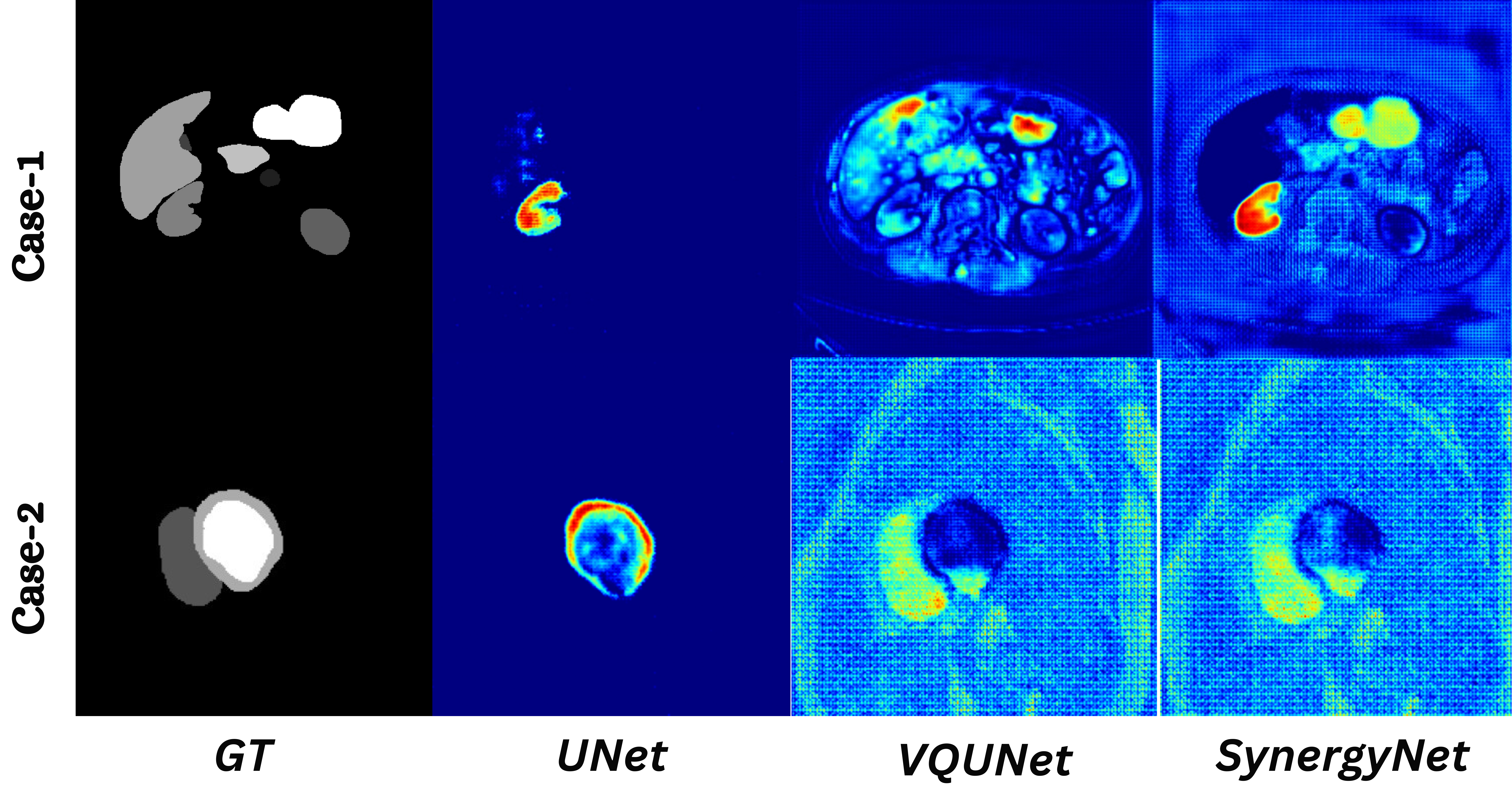}
\caption{Grad-CAM visualization.}
\label{fig: featuremapviz}
\vspace{-1mm}
\end{figure}

\textbf{Interpretability Analysis:} Here, we analyze the bottleneck architecture to evaluate learned representations. Fig.~\ref{fig: featuremapviz} visualizes the GradCAMs, revealing that CLS methods excel in capturing fine organ boundaries, while DLS methods excel in locating organs but struggle with fine boundary details. In contrast, SynergyNet effectively captures both fine and coarse boundaries, emphasizing the importance of leveraging complementary information. These findings further support the significance of synergistic effects.

\subsection{Cardiac Segmentation}
From Table~\ref{tab:cardiac_quantitative}, we note that the proposed SynergyNet outperforms both continuous and discrete baselines. SynergyNet can effectively capture complex heterogeneous structures. Compared to TransUNet and TransVQUNet-8s2h, SynergyNet-8s2h demonstrates 0.07pp and 11.61pp higher DSC and 0.06pp and 3.23pp lower HD. The qualitative results are shown in Fig.~\ref{pred:synaps_qualitative} further validate effectiveness of our approach in delivering more accurate segmentation results. 

\begin{table}[htbp]
\centering
\caption{Quantitaive results for cardiac segmentation}
\label{tab:cardiac_quantitative}
\adjustbox{width=0.48\textwidth}{
\begin{tabular}{lcc|ccc}
\toprule
\textbf{Method} & \multicolumn{2}{c}{\textbf{Mean Scores}} & \multicolumn{3}{c}{\textbf{Class-wise DSC}} \\
\cmidrule(lr){2-3} \cmidrule(lr){4-6}

\textbf{} & \textbf{DSC} & \textbf{HD} & \textbf{RV} & \textbf{Myo} & \textbf{LV} \\

\midrule
R50 UNet     & 87.94  &    2.01     & 84.62  & 
84.52  & 93.68 \\

DeepLabV3+     & 88.35  &    4.45     & 85.65  & 85.55  & 93.85 \\

R50 AttnUNet   & 86.90  &    2.10    & 83.27  & 84.33  & 93.53 \\

R50 ViT        & 86.19  &    1.98    & 82.51  & 83.01  & 93.05 \\
TransUNet     & \textbfblue{89.71}  &    \textbfblue{1.92}    & 86.67  &  \textbfgreen{87.27}  &  \textbfgreen{95.18} \\
\midrule

VQUNet            &     78.15    &    3.19     &    70.14 &    74.13  &  90.13         \\
TransVQUNet-2s2h    &     {74.40}     &    4.27     &   {64.69}      &     {72.75}     &     {85.77}     \\
TransVQUNet-8s2h    &   78.17      &    {4.63}      &  69.48       &    75.95     &  89.05       \\
\midrule

SynergyNet-2s2h     &   88.96      &   2.41      &   {86.80}      &    85.51     &   94.60      \\
SynergyNet-8s2h      &  \textbfgreen{89.78}       &    \textbfgreen{1.86}      &     \textbfgreen{87.68}     &  \textbfblue{86.60}       &   \textbfblue{95.06}      \\
SynergyNet-8s8h     &   89.37      &   2.14      &   \textbfblue{87.63}      &    86.49     &   94.98      \\
\bottomrule
\end{tabular}}
\end{table}


\subsection{Skin Lesion Segmentation}

\begin{figure*}[!ht]\centering
\def\svgwidth{\columnwidth}
\includegraphics[width=1.8\columnwidth]{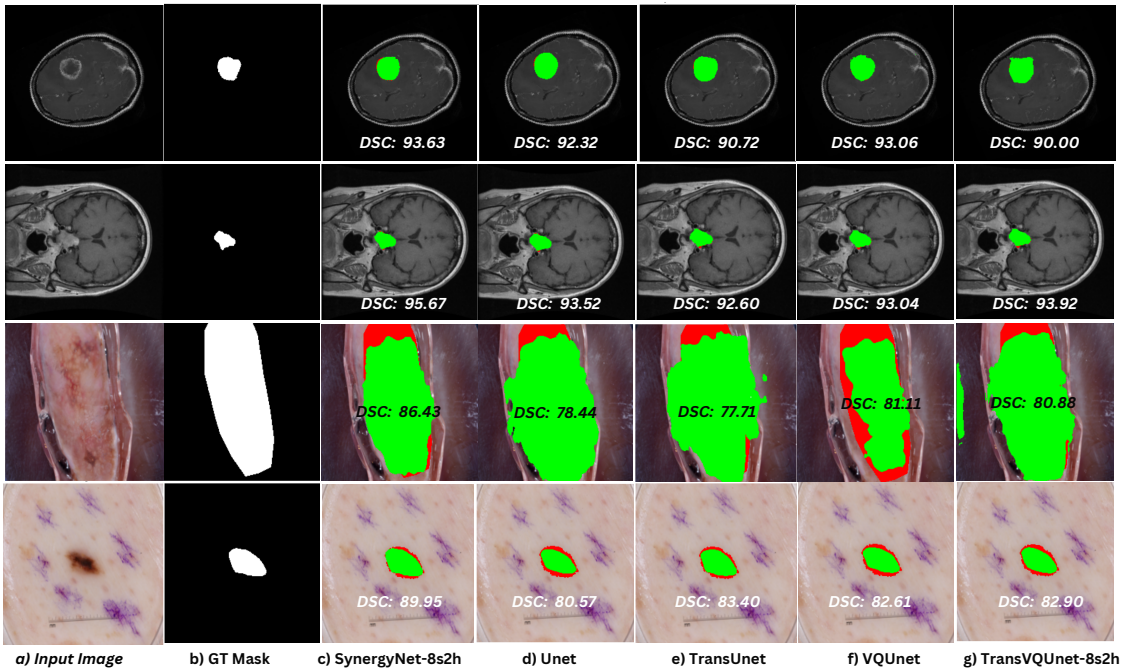} 
\caption{Segmentation maps on BTD (first two rows) and ISIC-2018 (last two rows) datasets. Actual and predicted pathological regions are shown in \textbfred{Red} and \textbfgreen{Green}, respectively.
}
\label{pred:skin_brain_preds}
\vspace{-5mm}
\end{figure*}

Table~\ref{tab:isic18} demonstrates the quantitative results on the ISIC 2018 dataset. Compared to CLS  methods, DLS-based approaches can effectively capture shapes like lesions, which typically exhibit less variability in terms of shape and size compared to organs and cardiac structures. However, the proposed SynergyNet method consistently outperforms both CLS and DLS-based methods, showcasing its ability to generalize well across different scenarios. Fig.~\ref{pred:skin_brain_preds} further highlights SynergyNet's ability to capture both coarse and fine-grained structured skin lesions. CLS-based methods tend to over-segment non-contour structures, while DLS-based methods such as VQUNet tend to under-segment lesions. In contrast, SynergyNet successfully and accurately segments lesions with smoother boundaries, demonstrating the importance of learning synergistic representations.

\begin{table}[htbp]
\centering
\caption{Quantitaive results for Skin lesion segmentation}
\label{tab:isic18}
\adjustbox{width=0.48\textwidth}{
\begin{tabular}{lccccc}
\toprule
\textbf{} & \textbf{IOU} & \textbf{DSC} & \textbf{ACC} & \textbf{SP} & \textbf{SE} \\
\midrule
R50 UNet     & 77.86 & 87.55 &  94.05 &  96.69  & 85.86 \\
DeepLabV3+  & 78.52 & 87.59 & 94.29 & 95.97 & 86.46 \\
R50 AttnUNet   & 78.43 & 87.91 & 94.13 & 96.23 & 87.60 \\ 
R50 ViT        & 78.13 & 87.45 & 93.53 & 96.13 & 87.10 \\ 
TransUNet     & 78.97 & 88.25 & 94.32 & 96.48 & 87.60 \\
\midrule

VQUNet & 79.13 & 88.35 & 94.46 & \textbfblue{97.09} & 86.29 \\
TransVQUNet-2s2h & 79.83 & {88.78} & {94.58} & 96.62 & \textbfblue{88.21} \\ 
TransVQUNet-8s2h & 79.68 & 88.69 & 94.54 & 96.64 & 88.01 \\
\midrule

SynergyNet-2s2h & {79.80} & 88.77 & 94.56 & 96.59 & \textbfgreen{88.26} \\ 
SynergyNet-8s2h & \textbfgreen{80.68} & \textbfgreen{89.31} & \textbfgreen{94.91} & \textbfgreen{97.37} & 87.28 \\ 

SynergyNet-8s8h & \textbfblue{80.05} & \textbfblue{88.92} & \textbfblue{94.66} & 96.81 & 87.98 \\ 
\bottomrule
\end{tabular}}
\end{table}

\subsection{Brain Tumour Segmentation}
SynergyNet achieves the best score on all metrics on the the BTS dataset (Table~\ref{tab:brain_tumour}) and outperforms the second-best method by a large margin. 
From Fig.~\ref{pred:skin_brain_preds}, we note that DLS methods tend to lose boundary information, but they  segment regions of interest more accurately than CLS methods for this particular case. On the other hand, SynergyNet consistently identifies regions of interest with smoother boundaries, surpassing both CLS and DLS methods. SynergyNet accurately predicts lesions, even in case of varying locations, sizes, and modality views. It effectively suppresses irrelevant information, such as the background.

\begin{table}[!t]
\centering
\caption{Quantitaive results for Brain tumour segmentation}
\label{tab:brain_tumour}
\adjustbox{width=0.48\textwidth}{
\begin{tabular}{lccccc}
\toprule
\textbf{Method} & \textbf{IOU} & \textbf{DSC} & \textbf{ACC} & \textbf{SP} & \textbf{SE} \\
\midrule
R50 UNet     & 63.90 & 78.00 & 99.33 & \textbfblue{99.77} & 77.22 \\
DeepLabV3+     & 65.23 & 78.90 & \textbfblue{99.60} & \textbfblue{99.77} & 77.22 \\
R50 AttnUNet   & 63.15 & 77.65 & 99.21 & 99.57 & 76.88  \\ 
R50 ViT        & 62.13 & 76.35 & 99.18 & 99.44 & 76.62 \\ 
TransUNet     & 62.36 & 76.82 & 99.24 & 99.62 & 76.90
 \\
\midrule
VQUNet & 58.00 & 73.40 & 99.11 & 99.66 & 72.66 \\
TransVQUNet-2s2h & 60.92 & 75.72 & 99.10 & 99.53 & 72.24 \\ 
TransVQUNet-8s2h & 61.22 & 75.94 & 99.25 & 99.73 & 72.44 \\
\midrule
SynergyNet-2s2h & \textbfblue{70.64} & \textbfblue{82.81} & {99.52} & 99.66 & \textbfblue{80.25} \\ 
SynergyNet-8s2h &  \textbfgreen{70.94} & \textbfgreen{83.00} & \textbfgreen{99.55} & \textbfgreen{99.86} & \textbfgreen{80.45} \\
SynergyNet-8s8h &  {69.88} & {82.27} & {99.43} & {99.76} & {79.85} \\
\bottomrule
\end{tabular}}
\end{table}

\vspace{-1mm}
\section{Ablation Studies}
Unless otherwise mentioned, we use $K=1024$, $dim=512$, $h_h=8$, $h_s=2$ and backbone as ResNet-50.

\begin{table}[htbp]
\centering
\caption{Codebook size (${K}$) analysis}\label{tab:codebook_analysis}
\begin{tabular}{cc|cc|cc}
\toprule
\multirow{2}{*}{\textbf{${K}$}} & \multirow{ 2}{*}{\textbf{${dim}$}} & \multicolumn{2}{c} {\textbf{Synapse}} & \multicolumn{2}{c}{\textbf{ACDC}} \\
\cmidrule(lr){3-4} \cmidrule(lr){5-6}
& & \textbf{DSC} & \textbf{HD} & \textbf{DSC} & \textbf{HD} \\
\midrule
1024 & 512 & \textbfgreen{79.65} & \textbfgreen{23.29} & \textbfgreen{89.78} & 1.86 \\
512 & 512 & 79.61 & 23.89  & 88.89 & 2.42 \\
256 & 512 & 79.21 & 29.47 & 89.29 & \textbfgreen{1.62} \\
128 & 512 & 78.48 & 30.97 & 89.01 & 1.79 \\
64 & 512 & {77.29} & {88.67} & {88.79} & {1.98} \\
\bottomrule
\end{tabular}
\end{table}

\subsection{Codebook analysis}\label{sec:codebook_analysis}
From Table~\ref{tab:codebook_analysis}, we observe a direct relationship between $K$ and the performance of SynergyNet on the Synapse dataset, where increasing $K$ leads to a notable improvement in HD scores.
On the ACDC dataset, the trend is different, such that $K=256$ gives the best HD score, and $K=128$ and $K=1024$ give comparable results. A smaller codebook size in the quantization module leads to higher compression and more aggressive quantization, but it can result in the loss of local information. This loss of fine-grained details and subtle variations can negatively impact the segmentation model's ability to capture intricate boundaries, leading to lower HD scores.  To achieve the best segmentation performance, the codebook size needs to be chosen so as to balance compression and preservation of local information.

\subsection{Hidden Dimension ($dim$) Analysis}
From Table \ref{tab:hidden_dim_analysis}, we observe that using a codebook size of $K$ = 1024 with $dim$ (hidden dimension size) greater than 512 or $dim$ less than 512 leads to a deterioration in performance. Empirically, we found that setting $K$ to be twice the value of $dim$ ($K$ = 2 * $dim$) yields the best performance. Thus, multiple parameters, including the dataset characteristics, influence the overall performance.

\begin{table}[htbp]
\centering
\caption{Hidden Dimension ($dim$) Analysis}
\label{tab:hidden_dim_analysis}
\begin{tabular}{cc|cc|cc}
\toprule
\multirow{2}{*}{\textbf{${K}$}} & \multirow{ 2}{*}{\textbf{${dim}$}} & \multicolumn{2}{c}{\textbf{Synapse}} & \multicolumn{2}{c}{\textbf{ACDC}} \\
\cmidrule(lr){3-4} \cmidrule(lr){5-6}
& & \textbf{DSC} & \textbf{HD} & \textbf{DSC} & \textbf{HD} \\
\midrule
1024 & 2048 & {78.48} & {30.28} & 88.58 & 2.33\\
1024 & 1024 & 77.61 & 29.53 & 88.64 & 2.12\\
1024 & 512 & \textbfgreen{79.65} & \textbfgreen{23.29} & \textbfgreen{89.78} & \textbfgreen{1.86}\\
1024 & 256 & 79.29 & 30.07 & 89.18 & 2.29\\
1024 & 128 & 78.98 & 35.81 & 88.87 & 2.60\\
\bottomrule
\end{tabular}
\end{table}

\subsection{Bottleneck Size Analysis}
Table~\ref{tab:bottleneck_size_analysis} presents the impact of the size of the DisconX module and the Refinement Module of SynergyNet on the overall segmentation performance. The combinations $h_s=8, h_{h}=0$ and $h_{s}=2, h_{h}=0$ denote the configurations without the refinement module.
We observe a significant deterioration in the overall performance when the refinement module is not utilized. 
For the Synapse dataset, the best value of DSC is obtained for $h_{s}=8, h_{h}=2$ and the best value of HD is obtained for $h_{s}=8, h_{h}=8$. For the ACDC dataset,  the combination $h_{s}=8, h_{h}=2$ results in the best values of DSC and HD. 
Overall, the optimal module size is dataset and task-dependent. It is crucial to consider these factors when determining the optimal sizes for the DisConX and refinement modules.

\begin{table}[htbp]
\caption{Bottleneck Size Analysis}
\label{tab:bottleneck_size_analysis}
\centering
\begin{tabular}{cc|cc|cc}
\toprule
 \multirow{ 2}{*}{\textbf{\textbf{$h_{s}$}}} & \multirow{ 2}{*}{\textbf{\textbf{$h_{h}$} }} & \multicolumn{2}{c}{\textbf{Synapse}} & \multicolumn{2}{c}{\textbf{ACDC}} \\
\cmidrule(lr){3-4} \cmidrule(lr){5-6}
& & \textbf{DSC} & \textbf{HD} & \textbf{DSC} & \textbf{HD} \\
\midrule
8 & 0 & \num{78.62} & \num{26.82} & \num{87.95} & \num{2.33} \\
8 & 2 & \textbfgreen{79.65} & {23.29} & \textbfgreen{89.78} & \textbfgreen{1.86} \\
2 & 8 & \num{78.45} & \num{25.19} & \num{88.96} & \num{2.41} \\
8 & 8 & \num{77.33} & \textbfgreen{20.56} & {89.68} & {2.14} \\

2 & 2 & \num{78.81} & \num{26.19} & \num{88.96} & \num{2.41} \\
2 & 0 & \num{78.12} & \num{27.71} & \num{87.36} & \num{3.11} \\
\bottomrule
\end{tabular}
\end{table}

\subsection{Contribution of DisConX Module}
The DisConX module plays a crucial role in the SynergyNet's ability to learn fine-grained local features. To understand  its contribution,  we create a variant SynergyNet(Fusion), which replaces the DisConX module with a simple feature fusion approach. It combines discrete and continuous representations and passes them through a refinement module. As shown in Table~\ref{tab:relevance_analysis}, this variant attains lower performance, which clearly demonstrates that the DisConX module is essential for learning fine-grained local features. Overall, results indicate that selectively attending to complementary information preserves higher-quality discriminative and semantic information.

\begin{table}[htbp]
\centering
\caption{Contribution of DisConX module}
\label{tab:relevance_analysis}
\begin{tabular}{c|cc|cc}
\toprule
\multirow{ 2}{*}{\textbf{Backbone}} & \multicolumn{2}{c} {\textbf{Synapse}} & \multicolumn{2}{c}{\textbf{ACDC}} \\
\cmidrule(lr){2-3} \cmidrule(lr){4-5}
&  \textbf{DSC} & \textbf{HD} & \textbf{DSC} & \textbf{HD} \\
\midrule

SynergyNet(Fusion) &  78.79	& 26.91 & 88.20 & 2.57\\
SynergyNet &  \textbfgreen{79.65} & \textbfgreen{23.29}  & \textbfgreen{89.03} & \textbfgreen{2.17}\\

\bottomrule
\end{tabular}
\end{table}

\subsection{Backbone Analysis}
We evaluate SynergyNet with ResNet and EfficientNet backbones. For the Synapse dataset, ResNet-50 achieved a DSC score of 79.65\%, and EfficientNet-B7 achieved the lowest HD score of 21.53\%. In the ACDC dataset, ResNet-101 performed the best on both metrics. EfficientNet-B0 exhibited remarkable boundary delineation capabilities despite its shallower architecture. Please Refer to the supplementary materials for parameters analysis.


\begin{table}[htbp]
\centering
\caption{Backbone Analysis}
\label{tab:backbone_analysis}
\begin{tabular}{c|cc|cc}
\toprule
\multirow{ 2}{*}{\textbf{Backbone}} & \multicolumn{2}{c} {\textbf{Synapse}} & \multicolumn{2}{c}{\textbf{ACDC}} \\
\cmidrule(lr){2-3} \cmidrule(lr){4-5}
&  \textbf{DSC} & \textbf{HD} & \textbf{DSC} & \textbf{HD} \\
\midrule
ResNet-18 &  77.28 & 27.88 & 88.82 & 2.08\\
ResNet-34 &  78.35 & 25.58 & 89.42 & 1.93\\
ResNet-50 & \textbfgreen{79.65} & 23.29 & 89.78 & 1.86\\
ResNet-101 & 78.66 & 28.50 & \textbfgreen{91.49} & \textbfgreen{0.91}\\
EfficientNet-B0 & 78.05 & 27.77 & 91.13 & 1.29 \\
EfficientNet-B7 & 78.70	& \textbfgreen{21.53} & 90.81 & 2.90\\
\bottomrule
\end{tabular}
\end{table}

\textbf{Limitations:} i) The quantizer's reliance on selecting the most similar codebook item for input representation may lead to difficulties in capturing intricate patterns, potentially causing information loss. ii) Both CLS and DLS struggle to effectively model inter-class relationships, resulting in increased false negatives. SynergyNet reduces false negatives compared to CLS and DLS but still can be further improved.






\section{Conclusion and Future Work}
We propose SynergyNet, a novel bottleneck architecture for learning complementary information from CLS and DLS. Extensive experiments and ablation studies confirm that SynergyNet captures both fine and coarse-grained details in the learned representations and outperforms previous works.  SynergyNet is a promising model for medical image analysis that offers high performance. Integrating SynergyNet with efficient architectures like Swin Transformer~\cite{cao2022swin}, HiFormer~\cite{heidari2023hiformer} and others shows promise for further advancements. Exploring SynergyNet's performance with unsupervised models is an intriguing research area that enables leveraging unlabeled data to enhance capabilities in medical image analysis. This holds the potential to improve efficiency and performance in this critical domain. 

{\small
\bibliographystyle{ieee_fullname}
\bibliography{egbib}
}

\end{document}